\documentclass[conference,9pt]{IEEEtran}
\IEEEoverridecommandlockouts

\usepackage{cite}
\usepackage{amsmath,amssymb,amsfonts}
\usepackage{algorithmic}
\usepackage{graphicx}
\usepackage{textcomp}
\usepackage{xcolor}
\usepackage{booktabs}
\usepackage{arydshln}
\usepackage{enumitem}
\usepackage{placeins}
\usepackage{multirow}
\usepackage{threeparttable}
\usepackage[hidelinks]{hyperref}

\def\BibTeX{{\rm B\kern-.05em{\sc i\kern-.025em b}\kern-.08em
    T\kern-.1667em\lower.7ex\hbox{E}\kern-.125emX}}

\begin{document}

\title{Faster IndexTTS-2: Accelerating and Streaming Autoregressive Zero-Shot Text-to-Speech Synthesis on GPUs}

\author{\IEEEauthorblockN{Muyang Du, Shuang Yu, Junjie Lai}
\IEEEauthorblockA{NVIDIA \\
Shanghai, China \\
\{myrond, shuangy, julienl\}@nvidia.com}
}

\maketitle

\begin{abstract}
Autoregressive text-to-speech models achieve strong naturalness but suffer from slow inference due to sequential token generation, limiting their deployment in production applications that require low latency. IndexTTS-2 is a state-of-the-art autoregressive TTS model consisting of a GPT, a flow-matching Diffusion Transformer, and a vocoder. Despite its high synthesis quality, its inference speed barely reaches real-time without streaming or batching support. We present Faster IndexTTS-2, which accelerates all neural network components of IndexTTS-2 for production deployment on GPUs using NVIDIA TensorRT and TensorRT-LLM. Faster IndexTTS-2 also enables streaming synthesis for latency-sensitive interactive applications, and batched inference across all components to maximize GPU utilization. Experiments on the Seed-TTS benchmark for both English and Chinese demonstrate up to 5.0$\times$ speedup on the autoregressive GPT and 3.6$\times$ end-to-end, with minimal degradation in word error rate, speaker similarity, and naturalness. Our methodology provides a practical reference for efficiently accelerating similar autoregressive speech models on GPUs.
\footnote{Audio samples: \url{https://faster-indextts-2.github.io}}
\end{abstract}


\section{Introduction}
\label{sec:intro}

Recent advances in large-scale speech generation have enabled zero-shot text-to-speech models to synthesize natural speech from only a short reference audio clip, supporting applications such as virtual assistants, video dubbing, and AI content creation. Current approaches can be broadly classified into autoregressive~\cite{indextts2, cosyvoice, cosyvoice2, fishspeech, sparktts, qwen3tts, fireredtts, s5tts} and non-autoregressive~\cite{f5tts, maskgct, e2tts, zipvoice, instantspeech} paradigms. Non-autoregressive models based on flow matching enable parallel generation with competitive quality. In contrast, autoregressive models achieve stronger naturalness through sequential token-by-token generation, but this sequential nature also results in slow inference, limiting their deployment in low-latency production applications. IndexTTS-2~\cite{indextts2} is a representative state-of-the-art autoregressive TTS model consisting of an autoregressive Text-to-Semantic module, a Semantic-to-Mel module based on flow matching, and a vocoder. Although it has high synthesis quality, the original IndexTTS-2 based on PyTorch processes one utterance at a time without streaming or batching support, with inference speed barely reaching real-time, making it difficult to deploy in production.

Recent inference optimization frameworks~\cite{tensorrt, trtllm, vllm, sglang, onnxruntime} have greatly accelerated neural network deployment. Among these, NVIDIA TensorRT~\cite{tensorrt} accelerates inference through graph optimization, operator fusion, and mixed-precision inference. TensorRT-LLM~\cite{trtllm} extends TensorRT for autoregressive large language models, providing optimized attention kernels, KV-cache management, and efficient sampling strategies. Standard TensorRT is well suited for feedforward models where the full computation graph is known at build time, while autoregressive models require TensorRT-LLM's specialized features for incremental token generation with dynamic sequence lengths. However, TensorRT-LLM is primarily designed for language models and requires additional adaptation for speech models.

In this paper, we present Faster IndexTTS-2, a fully accelerated IndexTTS-2 on GPUs for production deployment using TensorRT and TensorRT-LLM. Our main contributions are as follows:
\begin{itemize}[leftmargin=*,nosep,topsep=2pt]
\item We accelerate all neural network components of IndexTTS-2 using TensorRT and TensorRT-LLM, achieving up to 3.6$\times$ end-to-end speedup and up to 5.0$\times$ speedup on the autoregressive GPT.
\item We present a reusable methodology for adapting TensorRT-LLM to autoregressive TTS models, providing a practical reference for accelerating similar autoregressive speech models on GPUs.
\item We enable streaming synthesis with low Time-To-First-Audio (TTFA), crucial for latency-sensitive interactive applications.
\item We enable batched inference across all components to maximize GPU utilization, improving synthesis concurrency and throughput.
\item We systematically benchmark both inference efficiency and synthesis quality, providing practical insights for production deployment of autoregressive speech models.
\end{itemize}

\begin{table*}[!t]
  \centering
  \caption{Latency breakdown, overall performance, and synthesis quality of IndexTTS-2 vs.\ Faster IndexTTS-2 across precisions on the Seed-TTS test sets. All latencies in milliseconds (ms). W8A16 and W4A16 quantization are applied only to the GPT component.}
  \label{tab:latency}
  \resizebox{\textwidth}{!}{
  \renewcommand{\arraystretch}{1.1}
  \begin{tabular}{cc*{6}{c}*{6}{c}}
    \toprule
    \multirow{3}{*}{\textbf{}} &
    \multirow{3}{*}{\textbf{}} &
    \multicolumn{6}{c}{\textbf{Seed-TTS test-en}} &
    \multicolumn{6}{c}{\textbf{Seed-TTS test-zh}} \\
    \cmidrule(lr){3-8} \cmidrule(lr){9-14}
    & &
    \multicolumn{2}{c}{\textbf{IndexTTS-2 (PyTorch)}} &
    \multicolumn{4}{c}{\textbf{Faster IndexTTS-2 (TRT/TRT-LLM)}} &
    \multicolumn{2}{c}{\textbf{IndexTTS-2 (PyTorch)}} &
    \multicolumn{4}{c}{\textbf{Faster IndexTTS-2 (TRT/TRT-LLM)}} \\
    \cmidrule(lr){3-4} \cmidrule(lr){5-8} \cmidrule(lr){9-10} \cmidrule(lr){11-14}
    & &
    FP32 & FP16 & FP32 & FP16 & W8A16 & W4A16 &
    FP32 & FP16 & FP32 & FP16 & W8A16 & W4A16 \\
    \midrule
    \multirow{4}{*}{\shortstack{Latency\\Breakdown}}
      & Preprocessing$\downarrow$ & 115.6 & 150.2 & \textbf{42.4} & 50.0 & 49.2 & 49.8 & 96.4 & 113.7 & \textbf{29.0} & 35.9 & 36.9 & 36.7 \\
      & GPT$\downarrow$      & 2539.7 & 3137.6 & 676.5 & 506.6 & \textbf{479.6} & 476.3 & 3598.7 & 4432.7 & 1004.7 & 748.7 & \textbf{701.5} & 736.0 \\
      & DiT$\downarrow$      & 423.7 & 494.1 & 481.1 & \textbf{281.5} & 297.4 & 296.9 & 415.8 & 483.4 & 626.3 & \textbf{375.5} & 392.8 & 400.1 \\
      & Vocoder$\downarrow$  & 68.0 & 94.6 & 60.3 & \textbf{35.8} & 35.9 & 35.0 & 74.7 & 90.2 & 71.5 & \textbf{50.4} & 50.3 & 51.2 \\
    \midrule
    \multirow{2}{*}{Overall}
      & Latency$\downarrow$ & 3147.2 & 3876.8 & 1260.8 & 874.5 & \textbf{862.7} & 858.5 & 4185.8 & 5120.3 & 1732.1 & 1211.1 & \textbf{1182.0} & 1224.6 \\
      & RTF$\downarrow$         & 0.84 & 1.03 & 0.34 & 0.24 & \textbf{0.23} & 0.24 & 0.76 & 0.93 & 0.31 & 0.22 & \textbf{0.21} & 0.22 \\
    \midrule
    \multirow{3}{*}{Quality}
      & WER (\%)$\downarrow$ & 1.84 & \textbf{1.82} & 2.03 & 1.97 & 2.01 & 2.08 & \textbf{1.01} & 1.07 & 1.09 & 1.08 & 1.02 & 1.15 \\
      & SIM-o$\uparrow$        & \textbf{0.706} & 0.705 & 0.703 & 0.698 & 0.699 & 0.686 & \textbf{0.765} & 0.765 & 0.764 & 0.761 & 0.761 & 0.755 \\
      & UTMOS$\uparrow$      & \textbf{3.62} & 3.60 & 3.61 & 3.53 & 3.53 & 3.54 & \textbf{2.99} & 2.98 & 2.99 & 2.94 & 2.94 & 2.97 \\
    \bottomrule
  \end{tabular}
  }
\end{table*}

\section{System Design}
\label{sec:system}

\begin{figure}[!t]
  \centering
  \includegraphics[width=0.8\columnwidth]{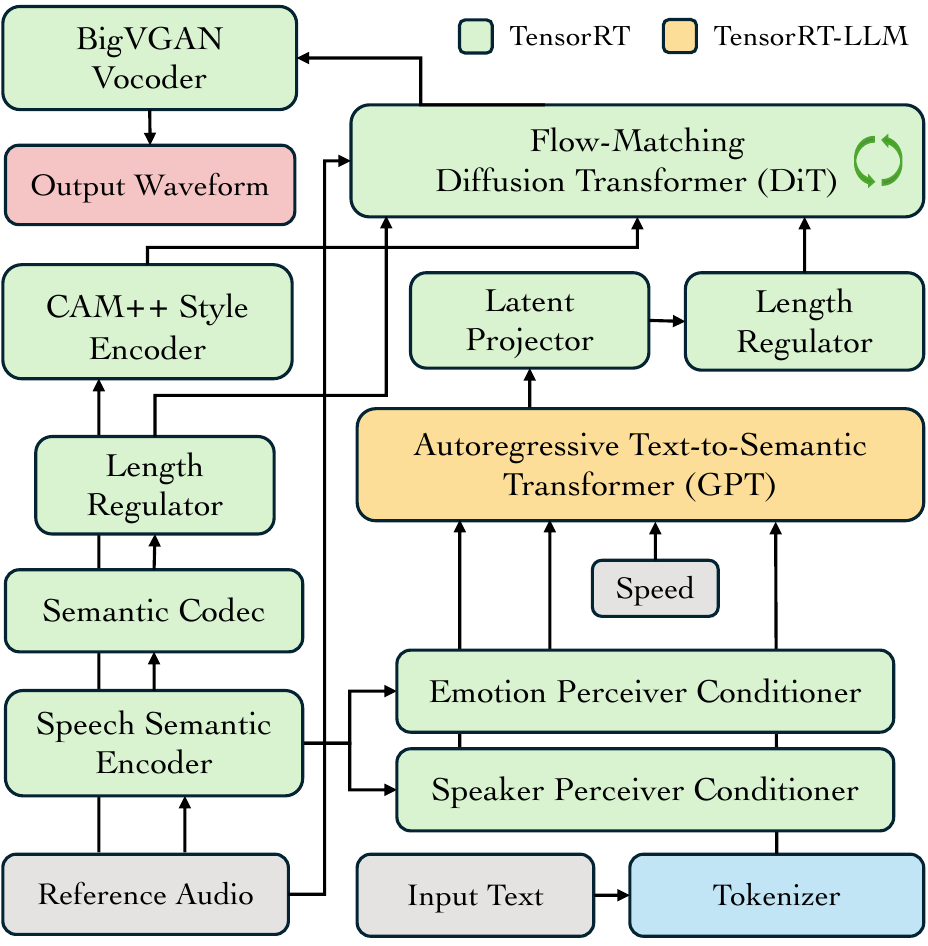}
  \caption{Overview of Faster IndexTTS-2.}
  \label{fig:overview}
\end{figure}

\subsection{Preliminary: IndexTTS-2}
\label{sec:preliminary}

IndexTTS-2~\cite{indextts2} is a cascaded autoregressive zero-shot text-to-speech model consisting of three core modules. The Text-to-Semantic module is a decoder-only autoregressive GPT~\cite{gpt2} that generates semantic codec tokens from text, conditioned on speaker latents and emotion embedding extracted from reference audio via Conformer-based~\cite{conformer} perceiver conditioners, along with a speed embedding. The Semantic-to-Mel module employs conditional flow matching~\cite{cfm} (CFM) with a Diffusion Transformer~\cite{dit} (DiT) to synthesize mel-spectrograms (Mel) from the semantic tokens, conditioned on the speaker prompt Mel, speaker style embedding, and GPT hidden states. At inference, an Euler ODE solver with classifier-free guidance~\cite{cfg} (CFG) iteratively refines the Mel over multiple steps. Finally, a BigVGAN~\cite{bigvgan} vocoder converts Mel to waveforms.

\subsection{System Overview}
\label{sec:sys_overview}

As illustrated in Figure~\ref{fig:overview}, Faster IndexTTS-2 divides the synthesis pipeline into four stages: preprocessing, GPT, DiT, and vocoder. The preprocessing stage encompasses a Wav2Vec2-BERT~\cite{w2vbert} semantic encoder, a VQ-VAE-based~\cite{vqvae} semantic codec adopted from MaskGCT~\cite{maskgct}, a CAM++~\cite{camplus} speaker style encoder, a length regulator, and speaker and emotion conditioners. The autoregressive GPT is accelerated with TensorRT-LLM and the remaining components are converted to TensorRT engines through a PyTorch-to-ONNX-to-TensorRT workflow with configurable precision and dynamic batch and sequence shapes. As a result, all neural network components in IndexTTS-2 are fully accelerated. Unlike the original IndexTTS-2, which only supports single-sample non-streaming inference, Faster IndexTTS-2 supports both streaming synthesis and batched inference across all stages with variable-length padding and masking, enabling concurrent synthesis of multiple utterances on a single GPU.

\subsection{GPT Acceleration with TensorRT-LLM}
\label{sec:gpt_accel}

TensorRT-LLM is originally designed for standard large language models and does not natively support the mixed input and output structure of IndexTTS-2's GPT. We apply several adaptations to bridge this gap. First, we leverage TensorRT-LLM's prompt tuning mechanism to inject conditioning signals. The conditioning embedding sequence, consisting of 32 speaker perceiver latents, 1 emotion embedding, and 1 speed embedding, is stored in the prompt embedding table and is retrieved via virtual input IDs prepended to the input sequence. Second, IndexTTS-2's GPT takes both text and semantic codec tokens as input, but TensorRT-LLM assumes a single input vocabulary with one embedding table. We then merge the text and codec embedding tables into a single table and shift text token IDs by the number of semantic codes at runtime to index the correct region. Third, we modify TensorRT-LLM to construct custom position IDs to match IndexTTS-2's behavior: conditioning latents receive empty positional embeddings, while text tokens and semantic tokens each use their own range of the single merged position embedding table. Fourth, standard TensorRT-LLM only outputs IDs with logits, but IndexTTS-2 requires per-step hidden states to enhance the downstream DiT. We therefore register an additional output tensor in the engine graph and patch TensorRT-LLM to support yielding the hidden states of the last GPT layer at each streaming generation step.

\subsection{Streaming Synthesis}
\label{sec:stream}

Faster IndexTTS-2 supports chunked streaming synthesis to reduce TTFA, also defined as the first-chunk latency. As the GPT generates semantic codec tokens autoregressively, chunks of a configurable size are decoded through the Semantic-to-Mel DiT and BigVGAN vocoder as soon as enough tokens accumulate, rather than waiting for full sequence completion. Adjacent chunks overlap by a configurable number of codec frames, and the overlapping waveform regions are blended using Hann window cross-fading to ensure smooth transitions. Batched streaming is also supported, where the GPT generates a batch of codec tokens at every step for multiple utterances and their codec chunks are converted to waveform chunks as a batch.

\section{Experiments}
\label{sec:experiments}

\begin{figure*}[!t]
  \centering
  \includegraphics[width=\textwidth]{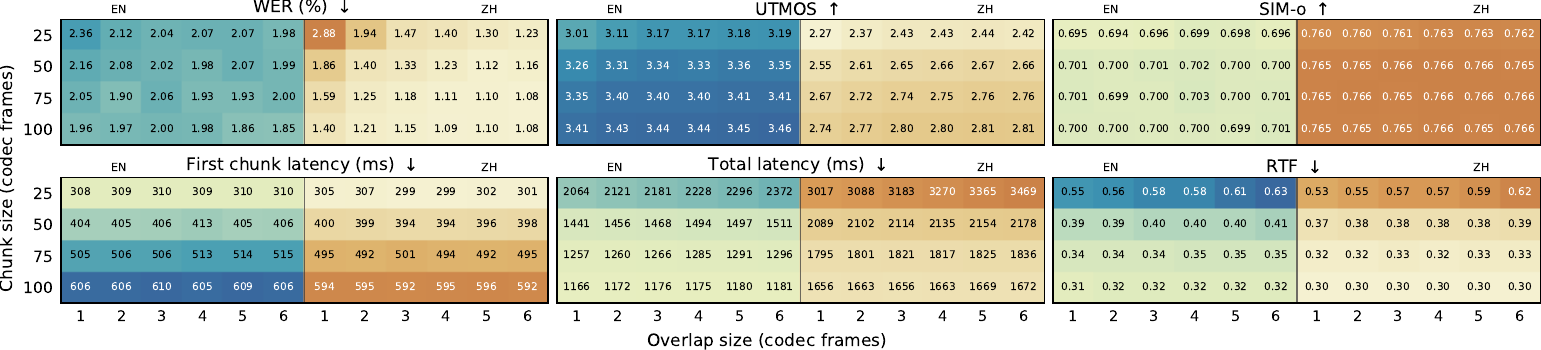}
  \caption{Streaming synthesis quality and latency of Faster IndexTTS-2 in FP16 across chunk sizes and overlap sizes on the Seed-TTS test sets.}
  \label{fig:streaming}
\end{figure*}

\subsection{Experimental Setup}
\label{sec:setup}

\noindent$\blacksquare$\ \textbf{Model.}\quad
Faster IndexTTS-2 has a total of 1.51B parameters: 788M in preprocessing, 512M in the autoregressive GPT, 96M in the flow-matching DiT, and 113M in the BigVGAN vocoder.

\noindent$\blacksquare$\ \textbf{Test Sets.}\quad
We evaluate on the Seed-TTS test sets~\cite{seedtts}, which contain 1,088 English samples sourced from Common Voice~\cite{commonvoice} and 2,020 Chinese samples from DiDiSpeech-2~\cite{didispeech}.

\noindent$\blacksquare$\ \textbf{Hardware and Software.}\quad
All experiments are conducted on a single NVIDIA A100 80GB Tensor Core GPU with an AMD EPYC 7J13 CPU. The software stack consists of Python 3.12, TensorRT 10.11, TensorRT-LLM 0.21, PyTorch 2.7.1, and CUDA 13.0.

\noindent$\blacksquare$\ \textbf{Metrics.}\quad
For quality, we measure word error rate (WER) using Whisper large-v3~\cite{whisper} for English and Paraformer-zh~\cite{funasr} for Chinese, speaker similarity (SIM-o) as the cosine similarity between reference and synthesized speaker embeddings from an ECAPA-TDNN~\cite{ecapatdnn} model trained on WavLM-large~\cite{wavlm} features, and speech naturalness via UTMOS~\cite{utmos}. For efficiency, we report total latency, real-time factor (RTF), time-to-first-audio (TTFA) for streaming, and throughput (Tput) in utterances-per-second for batched inference.

\subsection{Comparison of IndexTTS-2 and Faster IndexTTS-2}
\label{sec:latency}

As shown in Table~\ref{tab:latency}, Faster IndexTTS-2 in FP16 achieves a 3.60$\times$ end-to-end speedup on English and 3.46$\times$ on Chinese compared to IndexTTS-2 in PyTorch FP32, reducing the RTF from 0.84 and 0.76 to 0.24 and 0.22, respectively. The latency breakdown reveals that the autoregressive GPT dominates inference time, accounting for over 80\% of the total latency in the PyTorch baseline. TRT-LLM delivers the largest per-component speedup, accelerating GPT by 5.0$\times$ on English and 4.8$\times$ on Chinese and reducing its share to around 60\% after acceleration. The remaining components, preprocessing, DiT, and vocoder, are each accelerated 1.1 to 2.7$\times$ by TensorRT. Further weight-only quantizing the GPT to W8A16 or W4A16 yields only marginal additional latency reduction. Notably, IndexTTS-2 in PyTorch FP16 is over 20\% slower than in FP32, indicating that simply reducing precision in PyTorch does not guarantee faster inference.

In terms of synthesis quality, Faster IndexTTS-2 in FP16 preserves the quality of the original IndexTTS-2. WER increases by only 0.13 and 0.07 percentage points on English and Chinese, SIM-o decreases by less than 0.01, and UTMOS drops by less than 0.1 compared to the original IndexTTS-2 in FP32. Quantizing the GPT to W8A16 maintains quality comparable to FP16, while quantizing to W4A16 introduces slightly more quality degradation. Overall, both FP16 and W8A16 are suitable for production deployment, achieving 4 to 5$\times$ real-time synthesis with minimal quality loss.

\subsection{Performance of Streaming Synthesis}
\label{sec:streaming}

\begin{table}[!t]
  \centering
  \caption{Comparison of streaming and non-streaming for Faster IndexTTS-2 in FP16. All latencies in ms. Overlap is set to 5.}
  \label{tab:streaming}
  \resizebox{\columnwidth}{!}{
  \setlength{\tabcolsep}{2.7pt}
  \begin{tabular}{ccc*{3}{c}*{3}{c}}
    \toprule
    \multirow{2}{*}{\textbf{}} &
    \multirow{2}{*}{\textbf{Mode}} &
    \multirow{2}{*}[-2pt]{\textbf{\shortstack{Chunk\\(s)}}} &
    \multicolumn{3}{c}{\textbf{Efficiency}} &
    \multicolumn{3}{c}{\textbf{Quality}} \\
    \cmidrule(lr){4-6} \cmidrule(lr){7-9}
    & & &
    \textbf{TTFA$\downarrow$} & \textbf{Lat.$\downarrow$} & \textbf{RTF$\downarrow$} &
    \textbf{WER (\%)$\downarrow$} & \textbf{SIM-o$\uparrow$} & \textbf{UTMOS$\uparrow$} \\
    \midrule
    \multirow{3}{*}{EN}
      & NS & --- & --- & \textbf{874.5} & \textbf{0.24} & 1.97 & 0.698 & \textbf{3.53} \\
      & ST & 2 & 608.6 & 1180.4 & 0.32 & \textbf{1.86} & 0.699 & 3.45 \\
      & ST & 1 & \textbf{405.5} & 1496.7 & 0.40 & 2.07 & \textbf{0.700} & 3.36 \\
    \midrule
    \multirow{3}{*}{ZH}
      & NS & --- & --- & \textbf{1211.1} & \textbf{0.22} & \textbf{1.08} & 0.761 & \textbf{2.94} \\
      & ST & 2 & 596.1 & 1669.3 & 0.30 & 1.10 & 0.765 & 2.81 \\
      & ST & 1 & \textbf{395.9} & 2153.5 & 0.38 & 1.13 & \textbf{0.766} & 2.67 \\
    \bottomrule
  \end{tabular}
  }
\end{table}

Figure~\ref{fig:streaming} illustrates streaming synthesis quality and efficiency across chunk sizes and overlap sizes in GPT codec frames. Between the two parameters, chunk size is the primary factor affecting synthesis quality, as larger chunks yield lower WER and higher UTMOS on both English and Chinese test sets by reducing the number of chunk boundaries. Overlap further improves quality at small chunk sizes but has a diminishing effect as chunk size increases. SIM-o remains stable across all configurations, indicating that streaming does not affect speaker identity. While larger chunks increase TTFA, they reduce total latency (Lat.) and RTF since fewer chunks require fewer DiT and vocoder passes. Table~\ref{tab:streaming} compares non-streaming (NS) and streaming (ST) inference with an overlap of 5 and chunk sizes of 100 and 50, corresponding to 2s and 1s respectively. At the 2s chunk size, streaming preserves WER and SIM-o with UTMOS dropping by only 0.08 and 0.13 on English and Chinese. At the 1s chunk size, TTFA is lower at the cost of more UTMOS degradation and increased RTF. Despite some decrease in objective metrics, we observe barely perceptible difference in audio quality between streaming and non-streaming synthesis in practice. We encourage readers to listen to our audio samples for comparison.

\subsection{Performance of Batched Synthesis}
\label{sec:throughput}

\begin{table}[!t]
  \centering
  \caption{Batched inference performance for Faster IndexTTS-2 in FP16. Streaming chunk size is 2000\,ms with overlap of 5 frames. Tput\,=\,throughput (utterances/sec). All latencies in ms.}
  \label{tab:throughput}
  \resizebox{\columnwidth}{!}{
  \begin{tabular}{c*{3}{c}*{4}{c}}
    \toprule
    \multirow{2}{*}{\textbf{BS}} &
    \multicolumn{3}{c}{\textbf{Non-streaming}} &
    \multicolumn{4}{c}{\textbf{Streaming}} \\
    \cmidrule(lr){2-4} \cmidrule(lr){5-8}
    &
    \textbf{Lat.$\downarrow$} & \textbf{Tput$\uparrow$} & \textbf{RTF$\downarrow$} &
    \textbf{TTFA$\downarrow$} & \textbf{Lat.$\downarrow$} & \textbf{Tput$\uparrow$} & \textbf{RTF$\downarrow$} \\
    \midrule
    1  & 1097.0 & 0.912 & 0.2232 & 598.0 & 1492.9 & 0.670 & 0.3029 \\
    2  & 1595.7 & 1.253 & 0.1623 & 832.1 & 2238.6 & 0.893 & 0.2270 \\
    4  & 2657.9 & 1.505 & 0.1346 & 1328.2 & 3872.3 & 1.033 & 0.1956 \\
    8  & 4884.9 & 1.638 & 0.1233 & 2400.3 & 7441.9 & \textbf{1.075} & \textbf{0.1874} \\
    16 & 9574.8 & \textbf{1.671} & \textbf{0.1196} & 4694.1 & 15581.5 & 1.027 & 0.1940 \\
    \bottomrule
  \end{tabular}
  }
\end{table}

Table~\ref{tab:throughput} reports batched inference performance. Overall, batching improves throughput and batched RTF for both modes, with gains diminishing beyond batch size 8. Non-streaming shows only marginal improvement while streaming throughput slightly decreases and RTF slightly increases at batch size 16 due to repeated per-chunk overhead in the DiT and BigVGAN. Streaming TTFA increases with batch size, making large-batch streaming impractical for latency-sensitive scenarios. In deployment, non-streaming batching is suited for offline generation where throughput is prioritized, while streaming with small batch sizes serves applications requiring low first-chunk latency.

\section{Conclusion}
\label{sec:conclusion}

In this paper, we present Faster IndexTTS-2, a fully GPU-accelerated system for production deployment of IndexTTS-2 using TensorRT and TensorRT-LLM. By accelerating all neural network components and extending TensorRT-LLM with speech-specific adaptations for IndexTTS-2's autoregressive GPT, Faster IndexTTS-2 substantially reduces inference latency while preserving synthesis quality. We further enable streaming synthesis for latency-sensitive interactive applications and batched inference to maximize GPU utilization and throughput. Our systematic evaluation across different languages, precisions, streaming configurations, and batch sizes provides practical guidance for deploying autoregressive TTS models in production. The methodology we present for adapting TensorRT-LLM to TTS models is broadly applicable to similar autoregressive speech generation systems, and we hope this work encourages further research on efficient inference for speech models.

\section*{Acknowledgments}
Generative AI tools were used for grammar checking and language refinement to improve the clarity of the manuscript.

\newpage
\bibliographystyle{IEEEtran}
\bibliography{refs}

\end{document}